\documentclass[letterpaper, 10pt, conference]{ieeeconf}

\IEEEoverridecommandlockouts
\usepackage[utf8]{inputenc}
\usepackage[english]{babel}
\usepackage{amsmath}       
\usepackage{booktabs}       
\usepackage{array}         
\usepackage{graphicx}       
\usepackage{microtype}      
\usepackage{xurl}           
\usepackage{cite}           
\usepackage{etoolbox}       

\usepackage[hidelinks]{hyperref}

\let\OLDthebibliography\thebibliography
\renewcommand\thebibliography[1]{%
  \OLDthebibliography{#1}%
  \setlength{\parskip}{0pt}%
  \setlength{\itemsep}{-1pt plus 1pt}%
  \footnotesize
}

\title{\LARGE \bf
Not Forgotten: Implementation and Evaluation of a Personalized Episodic Memory for the Humanoid Robot Head Kim
}

\author{Steve Aschenbrenner$^{1}$ and Marcel Heisler$^{1}$ and Thomas Sievers$^{2}$ and Christian Becker-Asano$^{1}$
\thanks{$^{1}$Stuttgart Media University, Germany
        {\tt\small \{sa121, heisler, becker-asano\}@hdm-stuttgart.de}
        $^{2}$University of Lübeck {\tt\small t.sievers@uni-luebeck.de}}%
}

\begin{document}

\maketitle
\thispagestyle{empty}
\pagestyle{empty}

\begin{abstract}
Social robots that rely on large language models for conversation are unable to retain information across sessions. This absence of memory violates social expectations, potentially preventing the formation of persistent relationships. This paper presents a lightweight episodic memory module that integrates vector-based semantic retrieval with an LLM-controlled dialog system, deployed on the humanoid robot head Kim. The module employs a hybrid scoring function combining cosine similarity with a memory strength metric to retrieve contextually relevant past interactions and inject them into the generation prompt. The system was evaluated in a within-subjects video-based online study ($N = 43$) using the Human-Robot Interaction Evaluation Scale (HRIES). Results show that episodic memory significantly increased perceived sociability ($d = 0.60$, $p < .001$), with the strongest effects on perceived trustworthiness ($d = 0.62$) and warmth ($d = 0.56$). Perceived disturbance remained unchanged ($d = 0.00$), indicating that the implemented approach to personalized recall did not trigger privacy-related discomfort or uncanny valley effects. These findings suggest that episodic memory serves as a social lubricant in embodied Human-Robot Interaction, enhancing relational quality without eliciting negative affective responses.
\end{abstract}

\section{Introduction}
\label{sec:introduction}

Social robots designed for companionship, therapy support, and everyday assistance require the ability to maintain persistent relationships with their users~\cite{matheus_longterm_2025, gasteiger_factors_2023}. Unlike task-oriented industrial systems, these agents operate under an implicit social contract: users who share personal information expect that information to be retained across interactions~\cite{ray_being_2019}. Yet, large language models (LLMs) that increasingly serve as the conversational backbone of such robots are stateless by design~\cite{yu_stateful_2025}. Since interactions are either reset across sessions or constrained by finite context windows, they produce responses that are coherent in isolation but socially discontinuous over time. For text-based chatbots, this limitation is an inconvenience. For embodied humanoid robots, whose physical form invites attributions of agency and social capability~\cite{spatola_perception_2021, li_benefit_2015}, statelessness constitutes a violation of the expectations that embodiment itself creates~\cite{leite_social_2013}.

From an engineering perspective, the problem of persistence has been addressed. Retrieval-Augmented Generation (RAG)~\cite{lewis_retrievalaugmented_2020} and derivative architectures such as MemGPT~\cite{packer_memgpt_2024} enable LLMs to query external databases and inject retrieved context into the generation prompt. Simulated environments have further demonstrated that LLM-driven agents equipped with memory can exhibit believable social behaviors~\cite{park_generative_2023}. However, these systems are optimized for factual retrieval accuracy and evaluated primarily through functional benchmarks or virtual sandboxes. The transfer to embodied social robotics introduces a distinct concern that existing work has not adequately addressed: the impact of memory on human users.

This gap is twofold. First, the few empirical studies on memory in human-robot interaction (HRI) focus on system capability rather than user perception~\cite{pinto-bernal_designing_2025, mauliana_exploring_2025}. Research has concentrated on \textit{how to build}~\cite{peller-konrad_memory_2023} memory architectures rather than on \textit{how users experience} them. Whether episodic memory enhances perceived social attributes such as warmth, trustworthiness, or likeability of an embodied agent remains an open empirical question. Second, the introduction of personal recall by an artificial agent raises a complementary concern: the very act of a machine tracking and remembering past interactions may cross the boundary from attentive to intrusive~\cite{stein_venturing_2017, dorri_memory_2025}. It remains unclear whether users perceive a memory-enabled robot as a socially competent companion or if the retention of personal details triggers discomfort akin to surveillance. Determining whether memory-enabled robots are perceived as more sociable, more disturbing, or both requires controlled experimentation with human participants.

The central research question guiding this work is: \textit{What is the differential impact of a personalized episodic memory system on the perceived sociability and the perceived disturbance of a humanoid robot, compared to a Baseline system without memory?}

First, a lightweight episodic memory module is presented that integrates vector-based semantic retrieval via a Qdrant database with an LLM-controlled dialog system deployed on a humanoid robot head. Second, a controlled within-subjects study ($N = 43$) is reported that evaluates the effects of this memory system in a video-based paradigm using the validated Human-Robot Interaction Evaluation Scale (HRIES)~\cite{spatola_perception_2021}. Therefore, Section~\ref{sec:related_work} reviews existing memory architectures and their social implications. Section~\ref{sec:system_design} details the implementation of the proposed episodic memory module deployed on the robotic platform. 
Section~\ref{sec:study_design} outlines the methodology of the controlled user study, followed by the presentation of the empirical results in Section~\ref{sec:results}. Finally, Section~\ref{sec:discussion} discusses the findings along with their limitations, and Section~\ref{sec:conclusion} concludes with implications for future system designs.


\section{Related Work}
\label{sec:related_work}

\subsection{Episodic Memory in Social HRI}

The role of memory in sustaining long-term human-robot relationships has received growing attention. Leite et al.~\cite{leite_social_2013} identified long-term memory as a prerequisite for context-sensitive social interaction, arguing that agents lacking persistent recall cannot establish the continuity required for sustained user engagement. Peller-Konrad et al.~\cite{peller-konrad_memory_2023} provided a structured taxonomy of memory types for robotic agents, distinguishing procedural, semantic, and episodic subsystems. Within this hierarchy, episodic memory, defined as the encoding of specific past experiences including their temporal and contextual attributes~\cite{tulving_memory_1985}, is identified as the primary driver of perceived personalization. For instance, recent realizations of such social memories demonstrate that humanoid robots proactively adapt to their interaction partners by integrating dynamic person models with recognized emotional states~\cite{werk_how_2024}. Pinto-Bernal et al.~\cite{pinto-bernal_designing_2025} similarly reinforced this position, reporting that users who interacted with a memory-enabled social robot described the recall of personal details as creating a sense of continuity interpreted as social attentiveness. Their exploratory evaluation indicated that memory-driven personalization enhanced perceived trust and naturalness, constructs that relate to the broader social dimensions formalized in the Human-Robot Interaction Evaluation Scale (HRIES) by Spatola et al.~\cite{spatola_perception_2021}. However, these findings rest largely on qualitative observation rather than controlled experimental comparison.

Earlier foundational work by Kasap and Magnenat-Thalmann~\cite{kasap_building_2012} investigated the impact of episodic memory on an embodied, human-like robotic head in a tutoring scenario. In a controlled experiment comparing a memory-enabled versus a memoryless robot, they demonstrated that episodic memory significantly increased overall social presence and task engagement across multiple sessions. However, their dialogue management was heavily structured, relying on Finite State Machines (FSMs) and Hierarchical Task Networks (HTNs). Consequently, interactions were constrained to predefined, rule-based scripts, lacking the open-ended conversational flexibility and semantic adaptability that modern LLM-based architectures provide.

A complementary concern is that persistent recall may also increase user discomfort. Stein and Ohler~\cite{stein_venturing_2017} introduced the concept of the Uncanny Valley of Mind, proposing that attributed cognitive capabilities, including memory, can trigger eeriness when they exceed expected human-like boundaries. Dorri and Zwick~\cite{dorri_memory_2025} extended this argument by identifying a memory power asymmetry. The agent retains a complete record while the user naturally forgets, creating a dynamic that may be perceived as surveillance rather than social attentiveness.

\subsection{LLM-Based Memory Architectures}

The statelessness of LLMs has been addressed through external memory mechanisms. RAG~\cite{lewis_retrievalaugmented_2020} provides the foundational pattern. Relevant documents are retrieved from an external store and injected into the generation prompt. MemGPT~\cite{packer_memgpt_2024} extends this pattern with an explicit memory management layer that handles context overflow by swapping information between a limited working memory and a persistent long-term store. Both approaches optimize for factual retrieval accuracy in text-based assistant settings.

Park et al.~\cite{park_generative_2023} demonstrated that LLM-driven agents equipped with a memory stream, scored by recency, importance, and relevance, can maintain coherent social behavior in a simulated 2D environment over extended periods. Their Generative Agents architecture showed that appropriate memory retrieval directly contributed to perceived believability. Middleware frameworks such as LangChain~\cite{chase_langchain_2022} have since standardized the engineering components (vector stores, retrieval chains) required to implement such systems. Nevertheless, these architectures were largely developed and evaluated for disembodied or virtual agents, their optimization targets remaining closely tied to information accuracy rather than the social-perceptual quality of the interaction. Conversely, alternative approaches specific to HRI attempt to solve this by coupling LLMs with psychologically grounded cognitive architectures like ACT-R, through which humanoid robots retrieve memories via associative cognitive processes rather than pure semantic similarity~\cite{sievers_retrieving_2025}. While such neuro-symbolic systems offer excellent theoretical plausibility, evaluating the explicit psychological impact of any memory mechanism on human users during physical interaction remains critical.

\subsection{Embodiment and Memory Interdependence}

Physical embodiment fundamentally alters user expectations. Embodied agents elicit stronger attributions of agency and social capability than screen-based interfaces~\cite{li_benefit_2015, wainer_role_2006}, but this elevated expectation creates a liability. When an embodied humanoid fails to recall prior interactions, the resulting disappointment may reduce perceived social competence below that of a less anthropomorphic system~\cite{duffy_anthropomorphism_2003, reimann_survey_2024}. Conversely, memory without embodiment lacks the social presence needed to leverage recalled information for relationship building~\cite{lee_are_2006}. The combination of physical presence and memory-based consistency is therefore required for an agent to be perceived as a genuine social entity~\cite{leite_social_2013, mauliana_exploring_2025}.

\subsection{Research Gap}

The reviewed literature reveals a consistent pattern. Memory architectures have been built and evaluated in simulation~\cite{park_generative_2023} or assessed through technical retrieval metrics~\cite{packer_memgpt_2024, lewis_retrievalaugmented_2020}. In contrast, theoretical work in HRI has articulated the expected benefits of memory for sustained engagement~\cite{pinto-bernal_designing_2025, leite_social_2013} and provided structured architectural taxonomies~\cite{peller-konrad_memory_2023}, but without controlled empirical validation on embodied platforms. No existing study provides a controlled within-subjects comparison of a memory-enabled versus a memoryless agent in an embodied HRI context using a validated instrument. As a first empirical step, this paper deploys a RAG-based episodic memory module on a humanoid robot head and measures its effect on perceived sociability and disturbance through a video-based evaluation using the HRIES~\cite{spatola_perception_2021}.

\section{System Design}
\label{sec:system_design}

\begin{figure}[ht]
    \centering
    \includegraphics[width=0.5\linewidth]{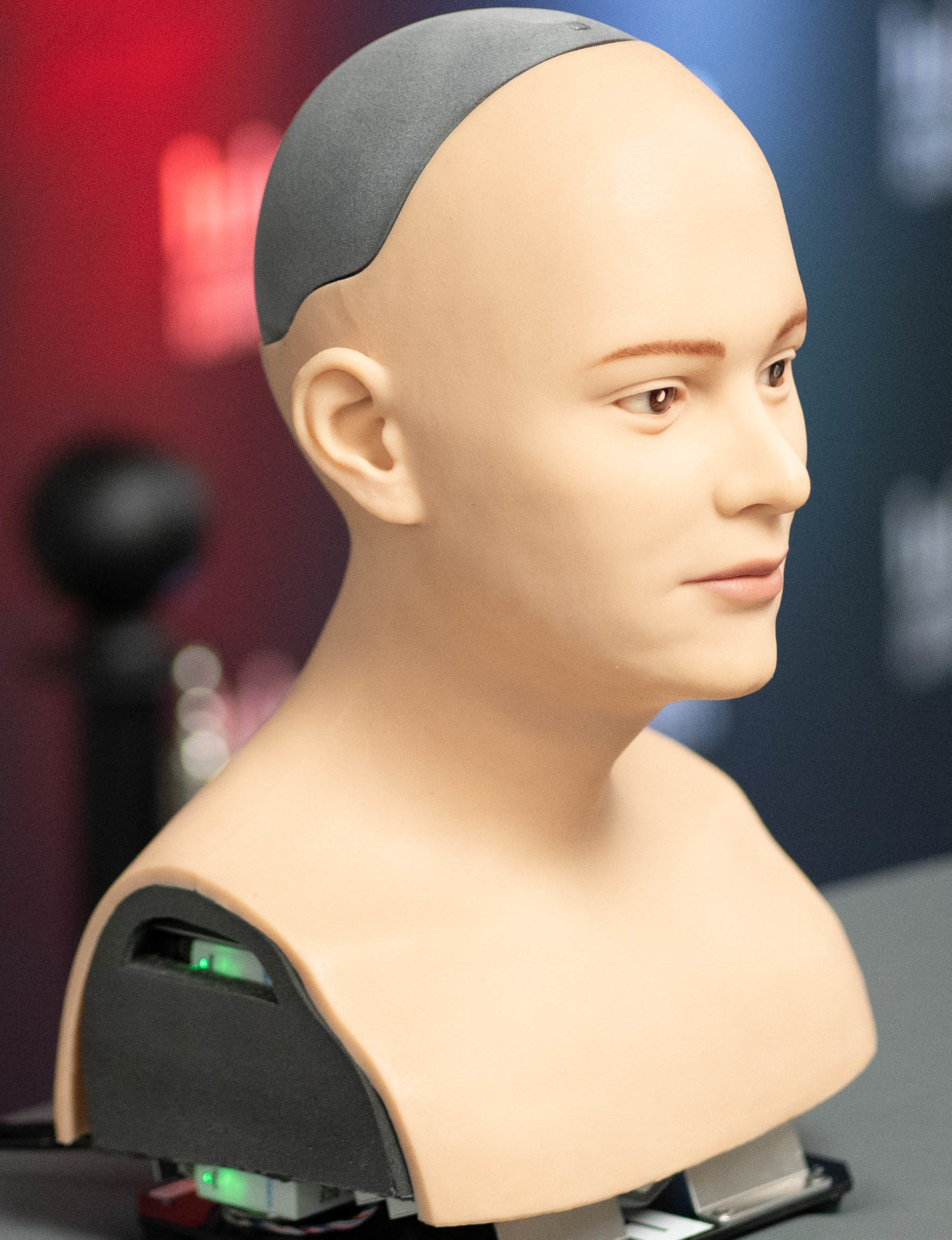}
    \caption{Humanoid robot head Kim that has been used in this study}
    \label{fig:robot-head-kim}
\end{figure}

\subsection{Architecture Overview and Robotic Platform}

\begin{figure*}[t]
  \centering
  \includegraphics[width=0.8\textwidth]{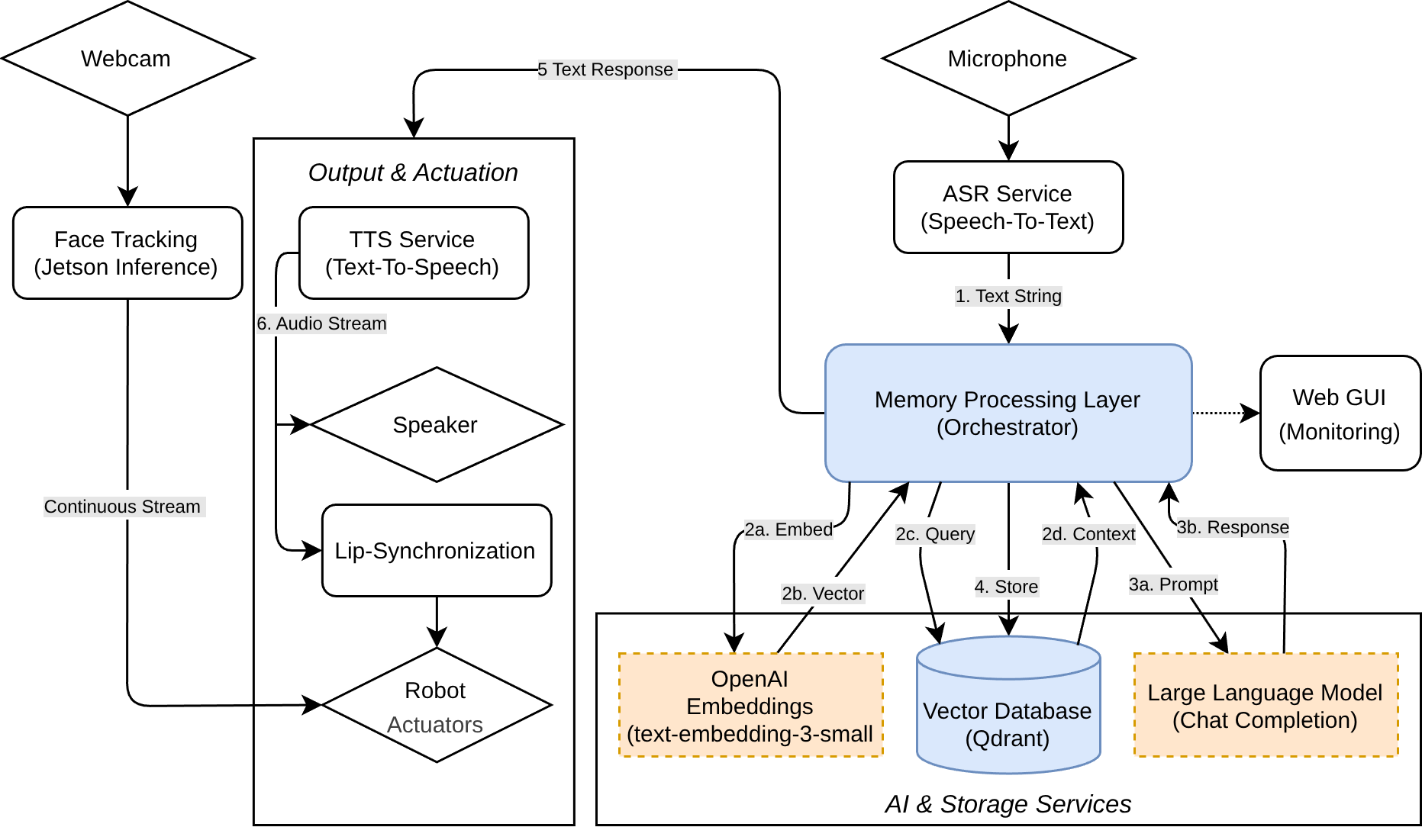}
  \caption{High-level system architecture. User speech is transcribed, enriched with retrieved episodic context, and passed to the LLM. Storage of new memories runs concurrently in a background thread.}
  \label{fig:architecture}
\end{figure*}

The system extends an existing robotic base infrastructure~\cite{heisler_android_2023} with a dedicated episodic memory module. The overall architecture follows a microservice-oriented design in which all components are containerized using Docker and communicate via a combination of HTTP REST APIs and websockets. The robotic base platform provides Automatic Speech Recognition (ASR) based on Whisper, Text-To-Speech (TTS) synthesis using XTTS~\cite{casanova_xtts_2024}, lip synchronization, and face tracking, each running as an isolated service on an Nvidia Jetson Orin AGX development board. The physical embodiment is a humanoid robot head (``Kim'') that is equipped with 14 pneumatic actuators for head movement and facial expression control~\cite{heisler_making_2023}, see Figure~\ref{fig:robot-head-kim}.

The interaction pipeline proceeds sequentially. Audio input is captured via microphone, transcribed by the ASR service, routed through the memory module for context retrieval, and forwarded to the LLM for response generation. The generated text is then synthesized into speech and rendered with synchronized lip movements. Figure~\ref{fig:architecture} illustrates the complete data flow, highlighting the memory module as an intermediary between perception and generation.

\subsection{Episodic Memory Module}

The memory module is the primary contribution of this work. Implemented as a FastAPI microservice, it exposes RESTful endpoints for storing new interactions (\texttt{POST /memory}), retrieving relevant past context (\texttt{POST /memory/query}), and triggering memory consolidation (\texttt{POST /memory/consolidate-all}). The system uses \texttt{gpt-4.1-nano} as its underlying language model, governed by a system prompt that explicitly defines its persona as an ironic, friendly android robot head named Kim.

\subsubsection{Vectorization and Storage}
Text fragments from each conversational turn are converted into 1536-dimensional vector embeddings using the OpenAI Embeddings API (\texttt{text-embedding-3-small}) and stored in a local Qdrant database. Vector storage facilitates retrieval by conceptual similarity rather than rigid keyword matching. While the architecture is designed to capture true episodic memories preserving temporal context, the experimental setup utilized pre-populated user preferences to ensure baseline consistency. Within the scope of this evaluation, the stored data thus functions as semantic profiles embedded within an episodic framework.

Each memory is additionally annotated with emotional metadata based on Russell's Circumplex Model~\cite{russell_circumplex_1980}. This psychological model represents emotions in a two-dimensional space defined by valence (pleasure-displeasure) and arousal (activation-deactivation). Valence $v$ and arousal $a$ scores are obtained through LLM-based sentiment classification at storage time. To achieve this, the model is explicitly instructed with the following prompt: 'EMOTION OUTPUT: Analyze the USER's emotional state (not your response). Valence: -1 (negative) to +1 (positive). Arousal: -1 (calm/sad/tired) to +1 (excited/angry/alert). Sad news like death = negative arousal. {"response": "text", "valence": float, "arousal": float}.' Based on this output, the emotional intensity is computed as $I = \sqrt{v^2 + a^2},$ (the Euclidean distance from the neutral origin in the Valence-Arousal plane). This emotional annotation enables the system to prioritize emotional events during retrieval.

\subsubsection{Hybrid Retrieval Scoring}
Retrieval follows a two-stage strategy. In the first stage, the system queries Qdrant using cosine similarity and over-fetches by a factor of three (i.e., $3k$ candidates for $k$ desired results). In the second stage, a hybrid scoring function re-ranks the candidates:

\begin{equation}
  R(m) = \alpha \cdot sim(m) + S(m),
  \label{eq:hybrid_score}
\end{equation}

\noindent where $R(m)$ is the final hybrid retrieval score, $sim(m) \in [0,1]$ is the cosine similarity score, $\alpha = 50$ is a heuristically calibrated scaling factor, and $S(m)$ is a memory strength score. While standard Retrieval-Augmented Generation architectures rely solely on semantic similarity, this approach adapts the recency and access frequency scoring from generative agents~\cite{park_generative_2023}. This approach is further extended by integrating emotional intensity, hypothesizing that prioritizing emotional events returns more natural recall. The scaling factor ensures semantic relevance remains the dominant criterion while memory strength acts as a tie-breaker.

The memory strength term is defined as

\begin{equation}
\begin{aligned}
S(m)
&=\max\!\Bigl(0,\Bigl( w_t F_{\text{time}} + w_a F_{\text{access}} \\
&\qquad +\; w_r F_{\text{recency}} \Bigr)\cdot M_{\text{emotion}}\Bigr),
\end{aligned}
\end{equation}

with weights $w_t=3.0$, $w_a=2.0$, and $w_r=1.5$. Its components are

\begin{equation}
\begin{aligned}
F_{\text{time}} &= \max\!\left(0,1-0.1\,\Delta t_{\text{creation}}\right),\\
F_{\text{access}} &= \ln\!\left(1+N_{\text{access}}\right),\\
F_{\text{recency}} &= \max\!\left(0,1-0.05\,\Delta t_{\text{last\_access}}\right),
\end{aligned}
\end{equation}

and

\begin{equation}
M_{\text{emotion}} = 1 + 0.5\,\sqrt{v^2+a^2},
\end{equation}

where $(v,a)$ are valence and arousal scores from Russell's Circumplex Model~\cite{russell_circumplex_1980}. Thus, emotional intensity acts as a multiplicative reinforcement, rendering highly emotional memories more resistant to decay.

Substituting these definitions yields the exact closed-form expression used in the implementation:

\begin{equation}
\begin{aligned}
S(m)=\max\!\Bigl(0,\Bigl(&3.0\cdot\max\!(0,1-0.1\,\Delta t_{\text{creation}}) \\
&+\;2.0\cdot\ln(1+N_{\text{access}}) \\
&+\;1.5\cdot\max\!(0,1-0.05\,\Delta t_{\text{last\_access}})\Bigr) \\
&\cdot\Bigl(1+0.5\cdot\sqrt{v^2+a^2}\Bigr)\Bigr),
\end{aligned}
\end{equation}

where $\Delta t_{\text{creation}}$ and $\Delta t_{\text{last\_access}}$ are measured in days, and $N_{\text{access}}$ is the retrieval count. The inner $\max(\cdot)$ terms clip negative time and recency contributions to zero.

After re-ranking, the top $k$ memories are injected into the LLM system prompt as contextual background. The prompt explicitly enforces conversational rules (e.g., 'use memories naturally', 'don't enforce all info at once'), framing retrieved context as optional background knowledge rather than mandatory facts to avoid forced or intrusive references.

Because retrieval adds latency (embedding: $\sim$0.32\,s; search/re-ranking: $\sim$0.36\,s), the controller splits the pipeline into blocking retrieval and non-blocking storage. Retrieval runs synchronously before generation, while memory writing is offloaded to a background thread.

\subsection{Experimental Condition Toggle}

To enable controlled comparison, the experimental manipulation was managed via container orchestration. In the Baseline condition, the memory Docker container was stopped. This forced the client's retrieval request to time out after 5 seconds, causing the LLM to generate responses using solely the current conversational turn with the memory prompt omitted. In the Memory-Enhanced condition, the active module queried the Qdrant database, which was pre-populated with the individual personal preferences collected from the respective participant during the first survey.

\section{Study Design}
\label{sec:study_design}

\subsection{Design and Conditions}

The evaluation followed a within-subjects design. Each participant experienced both experimental conditions: the Memory-Enhanced system with active episodic retrieval and a Baseline system without memory utilization. The presentation order of the two conditions was randomized to mitigate order effects. This design enables direct pairwise comparison while controlling for individual differences in personality traits and prior attitudes toward robots.

The study was conducted as an asynchronous video-based online study administered via the university's self-hosted LimeSurvey instance. The video-based format was selected for three reasons. First, pre-recorded stimuli ensured strict control over lighting, audio quality, and the robot's movement timing, variables that are difficult to standardize in live HRI studies. Second, the absence of an examiner during evaluation reduced social desirability bias. Third, the asynchronous format circumvented potential technical disruptions (e.g., network latency, TTS failures) and enabled location-independent recruitment.

The procedure consisted of two phases. In Phase~1 (Profiling), participants first gave informed consent after having read about the purpose of the study for which the robot head Kim was shown as a static picture. Then they completed a structured questionnaire collecting six personal preferences (comfort food, current hobby, disliked music genre, preferred morning beverage, dream travel destination, and name). These responses were converted manually into memory entries and injected into the Qdrant database via the storage API. To generate the experimental stimuli, the robot's resulting interactions were recorded with a webcam from a fixed frontal perspective. The raw footage was subsequently edited using Adobe Premiere to remove extended pauses. The prompts were visually overlaid onto the video, and subtitles were added for the robot's spoken responses. A week later, in Phase~2 (Evaluation), each participant viewed two of these video sequences showing the robot responding to six identical social prompts. The attached supplementary video showcases these exact six dialogue examples to illustrate the behavioral differences between the two study conditions. In the Baseline condition, the robot produced generic, non-personalized responses (e.g., \textit{``How about ordering some pizza?''}). In the Memory-Enhanced condition, responses incorporated retrieved personal information (e.g., \textit{``How about ordering some Ramen? You mentioned it's your go-to comfort meal.''}). The Baseline video was identical for all participants; Memory-Enhanced videos were individualized. The presentation order of the two conditions was randomized per participant to mitigate order and carry-over effects. Item order within the questionnaires was randomized independently.

\subsection{Hypotheses}

Two primary hypotheses guided the analysis, directly aligned with the core contribution:

\begin{itemize}
  \item \textbf{H1 (Sociability):} The Memory-Enhanced condition will lead to higher ratings on the Sociability subscale (items: \textit{warm, likeable, trustworthy, friendly}) compared to the Baseline condition.
  \item \textbf{H4 (Disturbance):} A memory-enabled robot is perceived as more or less disturbing as compared to a memoryless robot. While memory can foster connection, the precise recall of personal information could be perceived as intrusive, potentially evoking the Uncanny Valley of Mind effect~\cite{stein_venturing_2017}.
\end{itemize}

\noindent Additionally, Agency (H2) and Animacy (H3) were assessed as secondary measures to characterize the breadth of memory's perceptual impact. Finally, a forced-choice question was included to assess participants' global overall preference (H$_\text{G}$) between the two systems on an exploratory basis.

\subsection{Measurement Instrument}

Perceived social attributes were measured using the HRIES~\cite{spatola_perception_2021}. The HRIES comprises 16 adjective items distributed across four subscales: Sociability (warm, likeable, trustworthy, friendly), Agency (self-reliant, rational, intentional, intelligent), Animacy (alive, natural, real, human-like), and Disturbance (creepy, scary, uncanny, weird). Each item is rated on a 7-point Likert scale ranging from 1 (\textit{strongly disagree}) to 7 (\textit{strongly agree}). Subscale scores were computed as the arithmetic mean of the four constituent items. The HRIES was selected over alternatives such as the Godspeed Questionnaire~\cite{bartneck_measurement_2009} because it measures interaction-based perception rather than appearance-based impressions, providing greater sensitivity to the cognitive manipulation under investigation.

\subsection{Participants and Statistical Analysis}

Of 52 participants who completed Phase~1, $N = 43$ provided complete responses in Phase~2 and were included in the final analysis. Participants were recruited through university mailing lists at Stuttgart Media University, resulting in a sample consisting predominantly of students. Specific demographic data (such as exact age, gender, or prior robot experience) were not collected during the procedure following the rule of data minimization as stipulated by the General Data Protection Regulation.

The normality of difference scores (Memory-Enhanced minus Baseline) was verified for all subscales using Shapiro-Wilk tests. Paired-samples $t$-tests were employed as the primary inferential method. To control the family-wise error rate across four simultaneous comparisons (H1--H4), a Bonferroni correction was applied, yielding an adjusted significance threshold of $\alpha_{\text{adj}} = .0125$. Effect sizes were quantified using Cohen's $d$~\cite{cohen_statistical_2013}. A chi-square goodness-of-fit test assessed the global preference distribution (H$_\text{G}$).

\section{Results}
\label{sec:results}

\subsection{Reliability}

Internal consistency was assessed using Cronbach's alpha for each HRIES subscale under both conditions. All subscales demonstrated acceptable to good reliability ($\alpha \geq .67$). Sociability yielded $\alpha = .72$ (Baseline) and $\alpha = .74$ (Memory-Enhanced). Agency showed the lowest consistency ($\alpha = .67$ in both conditions). Animacy ($\alpha = .87 / .92$) and Disturbance ($\alpha = .87 / .86$) exhibited high internal consistency. These values confirm that the subscale scores can be interpreted with sufficient confidence.

\subsection{H1: Sociability}

Participants rated the Memory-Enhanced system significantly higher on Sociability than the Baseline system. The difference scores were normally distributed (Shapiro-Wilk $W = .97$, $p = .289$). A paired-samples $t$-test confirmed the effect: $t(42) = -4.15$, $p < .001$ (Bonferroni-corrected $p < .001$), with a medium-to-large effect size ($d = 0.60$).

The item-level analysis, detailed in Table~\ref{tab:results}, reveals that this effect was not uniform across items. The gains were primarily driven by \textit{trustworthy} ($M_{\text{BL}} = 3.21$, $M_{\text{ME}} = 4.09$, $d = .62$) and \textit{warm} ($M_{\text{BL}} = 3.07$, $M_{\text{ME}} = 3.91$, $d = .56$), both showing medium-to-large effects with $p \leq .001$. The item \textit{likeable} increased significantly with a smaller effect ($d = .33$, $p = .022$). In contrast, \textit{friendly} showed no significant change ($d = .24$, $p = .114$), having already received high Baseline ratings ($M = 5.00$).

\begin{table}[t]
\centering
\caption{Paired $t$-test results for Sociability and Disturbance. Bonferroni-corrected threshold: $\alpha_{\text{adj}} = .0125$.}
\label{tab:results}
\scriptsize
\setlength{\tabcolsep}{3pt}
\begin{tabular}{@{}llccccc@{}}
\toprule
 & \textbf{Item} & \textbf{BL} $M(SD)$ & \textbf{ME} $M(SD)$ & $t(42)$ & $p$ & $d$ \\
\midrule
\rotatebox{90}{\textbf{Soc.}} & warm        & 3.07\,(1.37) & 3.91\,(1.62) & $-$3.56 & .001   & .56 \\
            & likeable    & 3.81\,(1.45) & 4.28\,(1.33) & $-$2.38 & .022   & .33 \\
            & trustworthy & 3.21\,(1.32) & 4.09\,(1.51) & $-$3.91 & \textless.001  & .62 \\
            & friendly    & 5.00\,(1.40) & 5.33\,(1.34) & $-$1.61 & .114   & .24 \\
            & \textit{Scale} & 3.77\,(1.02) & 4.40\,(1.09) & $-$4.15 & \textless.001 & .60 \\
\midrule
\rotatebox{90}{\textbf{Dist.}} & creepy  & 4.23\,(1.93) & 4.19\,(2.01) & 0.32  & .750 & .02 \\
            & scary   & 3.49\,(1.79) & 3.74\,(1.95) & $-$1.60 & .117 & .14 \\
            & uncanny & 4.44\,(1.80) & 4.60\,(1.71) & $-$1.10 & .279 & .09 \\
            & weird   & 4.72\,(1.52) & 4.37\,(1.57) & 1.41  & .164 & .23 \\
            & \textit{Scale} & 4.22\,(1.50) & 4.23\,(1.53) & $-$0.05 & .960 & .00 \\
\bottomrule
\end{tabular}
\end{table}

\subsection{H4: Disturbance}

The Memory-Enhanced system did not increase perceived disturbance. The two conditions produced near-identical mean ratings ($M_{\text{BL}} = 4.22$, $M_{\text{ME}} = 4.23$). A paired-samples $t$-test yielded no significant difference: $t(42) = -0.05$, $p = .960$ (Bonferroni-corrected $p = 1.000$), $d = 0.00$. As shown in Table~\ref{tab:results}, none of the four constituent items approached significance individually, and all item-level effect sizes remained negligible ($d < .25$).

\subsection{Secondary Measures: Agency and Animacy}

\begin{figure*}[t]
  \centering
  \includegraphics[width=\textwidth]{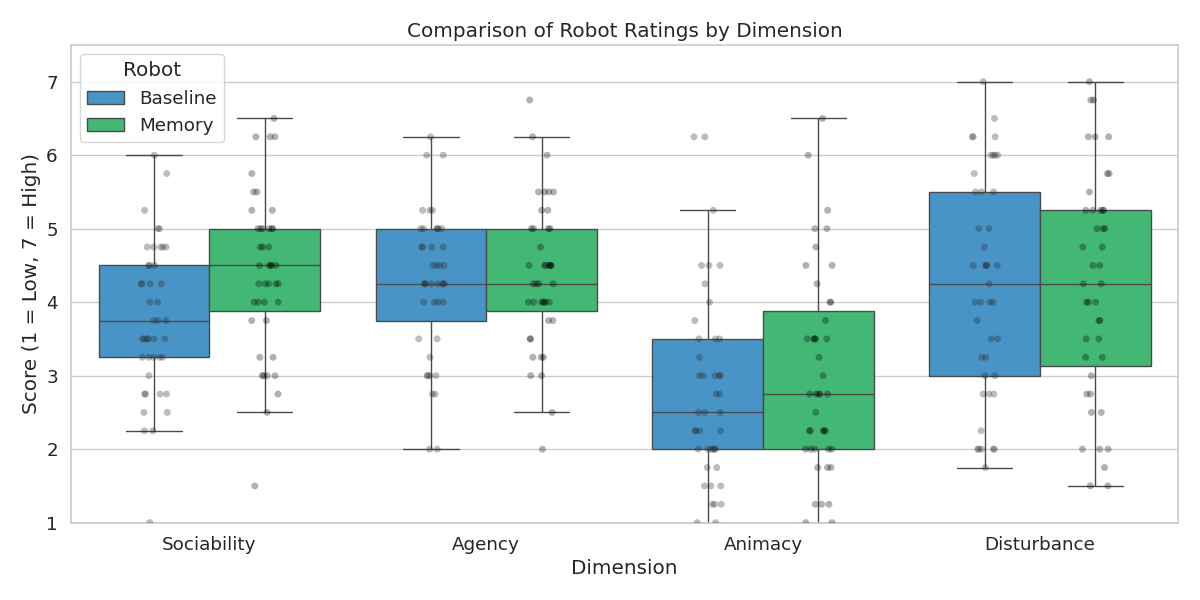}
  \caption{HRIES subscale scores by condition ($N = 43$). Only Sociability exhibits a visible distributional shift between Baseline (left) and Memory-Enhanced (right). The remaining subscales show largely overlapping distributions.}
  \label{fig:boxplots}
\end{figure*}

Neither Agency nor Animacy showed significant differences at the scale level. For Agency, the two conditions produced near-identical means ($M_{\text{BL}} = 4.25$, $M_{\text{ME}} = 4.36$; $t(42) = -0.89$, $p_{\text{corr}} = 1.000$, $d = 0.11$). For Animacy, a small non-significant trend was observed ($M_{\text{BL}} = 2.78$, $M_{\text{ME}} = 3.02$; $t(42) = -1.62$, $p_{\text{corr}} = .452$, $d = 0.17$), with both conditions remaining below the scale midpoint. Fig.~\ref{fig:boxplots} provides a visual comparison across all four subscales.

\subsection{Global Preference}

When asked to indicate an overall preference, 27 participants (63\%) selected the Memory-Enhanced system and 16 participants (37\%) selected the Baseline system. A chi-square goodness-of-fit test against an equal distribution yielded a non-significant result: $\chi^2(1) = 2.81$, $p = .093$, Cohen's $w = 0.26$. The observed preference distribution cannot be distinguished from chance at the conventional significance level.

\section{Discussion}
\label{sec:discussion}

\subsection{Episodic Memory as a Social Lubricant}

The significant sociability effect ($d = 0.60$) is consistent with three predicted theoretical mechanisms: the cumulative establishment of Common Ground~\cite{clark_grounding_1991, clark_common_2015}, the progression from superficial to intimate exchange described by Social Penetration Theory~\cite{altman_social_1973}, and the avoidance of a relational reset through the psychological significance of not being forgotten~\cite{ray_being_2019}.

The item-level pattern is informative. The effect was driven by \textit{trustworthy} ($d = .62$) and \textit{warm} ($d = .56$), while \textit{friendly} remained unchanged due to a ceiling effect ($M_{\text{BL}} = 5.00$). Basic politeness was attributed regardless of memory. What memory added was a deeper sense of being known. This aligns with Common Ground~\cite{clark_grounding_1991} and relational progression~\cite{altman_social_1973} affecting relational depth rather than courtesy. Trust requires an expectation of continuity which memory directly provides. Conversely, the Baseline system's failure to reference past exchanges communicated relational indifference~\cite{ray_being_2019}. This distinction offers a clear design guideline. Memory references are most valuable when they communicate care and attentiveness rather than merely demonstrating technical recall. Furthermore, while memory significantly enhanced these specific relational qualities, the global preference (63\% for Memory-Enhanced) did not reach significance. This suggests that improved social impressions do not automatically dictate overall preference, although the sample may have been underpowered to detect this exploratory measure.

These findings extend prior work by Leite et al.~\cite{leite_social_2013}, who identified the novelty effect as a core limitation of stateless architectures. The present results demonstrate that the mechanism of memory operates not merely through novelty preservation but through affective dimensions, trust and warmth, that were not the primary focus of earlier longitudinal studies.

\subsection{Absence of Disturbance}

The near-zero disturbance effect ($d = 0.00$) indicates that the memory system did not trigger the Uncanny Valley of Mind~\cite{stein_venturing_2017} or surveillance concerns associated with memory power asymmetry~\cite{dorri_memory_2025}. Two factors likely contributed. First, the hybrid scoring function (Eq.~\ref{eq:hybrid_score}), calibrated to prioritize semantic relevance ($\alpha = 50$) over access frequency, prevented the retrieval of irrelevant details. The prompt design further framed memories as optional background knowledge. This combination implemented the contextual selectivity that Grice's maxims of Quantity and Relation~\cite{grice_logic_1975} require for cooperative communication. Second, the stored preferences were relatively favorable and presented in a single video exposure. Participants might react differently if the robot recalled sensitive details unexpectedly during live interactions.
The present design cannot separate whether the absence of disturbance is caused by the architectural filtering or the limited scenario. Longitudinal deployments with organically growing databases are required to test this at scale. 

\subsection{Architectural Enablers}

Two architectural decisions appear to have been prerequisites for the observed effects. First, vector-based semantic retrieval, rather than keyword matching, enabled the system to bridge lexical paraphrases across sessions, producing topically appropriate memory references that supported the sense of shared understanding. Second, the hybrid scoring function ensured that the most semantically relevant memories were surfaced rather than the most frequently accessed ones, preventing irrelevant intrusions that could have disrupted both sociability gains and disturbance thresholds. The same filtering that prevented social violations, however, also concealed the selective cognition that could have enhanced perceived agency, an inherent trade-off in the current design.

\subsection{Scope of the Comparison}

The present study compared a memory-enabled system against a memoryless Baseline, establishing the perceptual value of personalization itself. This design does not isolate the contribution of the specific architectural decisions (vector-based semantic retrieval, hybrid scoring) from simpler alternatives such as keyword matching or template-based insertion. It is therefore possible that any system capable of injecting personal details into responses would produce a comparable sociability effect. Disentangling the contribution of the retrieval architecture from the mere presence of personalized content requires a multi-condition follow-up study comparing the proposed system against simpler memory implementations. The architectural claims in this paper are therefore limited to demonstrating feasibility and describing the mechanisms that enabled socially appropriate recall; the empirical contribution lies in the effect of memory availability, not in the superiority of a particular retrieval strategy.

\subsection{Limitations}

Several constraints bound the interpretation of these results. First, the video-based format excluded real-time dialogue and reduced the bandwidth through which memory could influence user judgments. Also, observing personalized responses in a single session limits ecological validity and prevents the empirical testing of long-term factors like time decay. The null results on Agency and Animacy may partly reflect this reduced validity~\cite{spatola_perception_2021}. Second, constructs such as Narrative Identity~\cite{mcadams_narrative_2011} and the Intentional Stance~\cite{dennett_intentional_2009} require sustained temporal depth to develop. Third, the sample of $N = 43$ university students limits statistical power. Furthermore, specific demographic data were not collected due to data minimization principles. This restricts the generalizability of the findings regarding individual differences in privacy attitudes. Fourth, LLM variability introduced uncontrolled within-group variance, as the quality of memory integration ranged from naturally woven references to scripted insertions.

\section{Conclusion and Future Work}
\label{sec:conclusion}

This paper investigates whether equipping a humanoid robot head with an LLM-integrated episodic memory module affects perceived social attributes. A lightweight memory architecture based on vector-based semantic retrieval and hybrid scoring is developed, and its impact is evaluated through a within-subjects study ($N=43$) using the HRIES. The central finding is that episodic memory significantly increases perceived sociability ($d=0.60$, $p<.001$), driven by gains in trust and warmth, while perceived disturbance remains unchanged ($d=0.00$). Memory operates as a social facilitator without eliciting privacy-related discomfort.

These results carry direct implications for HRI design. The observed effect of memory on sociability combined with the absence of effects on agency suggests that the architecture primarily influences relational perception. This highlights the potential of memory modules designed around social bonding features. The empirical results demonstrate that personalized recall improves sociability. However, the findings do not prove that the proposed hybrid architecture is superior to simpler methods like template-based keyword retrieval. The architectural claims validate viability rather than definitive superiority. Furthermore, it remains an open question whether a system with unfiltered recall would trigger privacy concerns. Designers should consider treating retrieval logic as a component with social-perceptual consequences, not merely as an information retrieval optimization problem.

Three directions for future work emerge from the identified limitations. First, a longitudinal study spanning multiple weeks with organically growing memory databases is required to test whether the null effects on Agency and Animacy emerge with greater interaction depth, and to empirically validate the forgetting mechanism. In this context, comparing the proposed LLM-driven filtering approach to the native cognitive decay mechanisms found in architectures like ACT-R could isolate how different algorithmic approaches to forgetting mitigate perceived disturbance. Second, live interaction studies should replicate and extend the present findings by restoring the real-time dialogue, non-verbal cues, and embodiment channels that the video-based format precludes. Third, future research should evaluate explainable memory retrieval strategies, where the agent occasionally verbalizes why it references a particular memory. Exposing this internal filtering process could unlock the agency gains that the current silent approach does not achieve.


\bibliographystyle{IEEEtran}
\bibliography{bibliography/masterthesis}

\end{document}